\documentclass[graybox]{svmult}

\usepackage{mathptmx}       
\usepackage{helvet}         
\usepackage{courier}        
\usepackage{type1cm}        
\usepackage{makeidx}         
\usepackage{graphicx}        
\usepackage{multicol}        
\usepackage[bottom]{footmisc}

\usepackage{booktabs}         
\usepackage{comment}         
\usepackage{url}
\usepackage[backend=bibtex,sorting=nyt,natbib=true,citestyle=authoryear-comp,bibstyle=authoryear,giveninits=true,maxbibnames=9]{biblatex}

\usepackage{color}

\bibliography{references}
 
\makeindex             

\begin{document}

\title*{Anchoring Distributed Agency: Institutional Metaphors for Designing Large-Scale AI and AI Techniques for Running Institutions}
\titlerunning{Anchoring Distributed Agency: Institutional Metaphors and AI Techniques} 
\author{Alexander Boer and Giovanni Sileno}
\institute{Alexander Boer \at KPMG, Amsterdam, the Netherlands. \email{boer.alexander@kpmg.nl}
\and Giovanni Sileno \at Socially Intelligent Artificial Systems, Informatics Institute, University of Amsterdam, Amsterdam, the Netherlands. \email{g.sileno@uva.nl}}
%
%
\maketitle

\abstract{Artificial Intelligence (AI) started out with the ambition to reproduce the human mind, but as soon as the sheer scale of that ambition became apparent, it retreated into either studying specialized intelligent behaviours or proposing overarching architectural concepts for interfacing specialized intelligent components---conceived of as agents acting through organizational structures. The agent-based modelling paradigm proved to have interesting applications beyond the original domain, supporting the understanding, simulation, and the assessment of behaviour of socio-economic and legal structures on an aggregate level. Looking at contemporary AI, whereas text completion provides a successful example of specialized behaviour, surprisingly little of the rich foundation of multi-agent systems (MAS) and normative systems research is currently reflected in Agentic AI systems. Addressing this gap, the chapter examines cross-cutting
conceptualisations, modelling problems, and design challenges shared by
large-scale distributed AI and institutional systems. Three main contributions can be highlighted.  First, by tracing the historical trajectory of AI, the chapter identifies an \textit{ecological} perspective on intelligence, examining both the value and limits of the \textit{compression-as-intelligence} hypothesis. Second, it shows how moving between internal cognitive mechanisms and external manifestations of agency supports a functional separation of concerns for both individual and collective agency. Third, it argues that \textit{epistemological}, \textit{normative}, and \textit{value pluralisms} are necessary conditions for sustaining the ecological character of intelligence. Furthermore, we indicate how AI techniques can support institutions by modelling social roles, simulating normative interactions, detecting and diagnosing failures, and supporting procedures of review and revision.}

\section{Introduction}\label{introduction}

Until recently, analogies between social institutions and computational mind architectures were largely theoretical design metaphors. We spoke of \textit{electronic institutions} as alternatives (or better, complementary) to traditional social arrangements, using institutional vocabulary---norms, roles, and obligations---to conceptualize distributed computational systems. Today, the rise of  Agentic AI, driven by large language models (LLMs), has made those theoretical analogies particularly relevant. As autonomous (or semi-autonomous) agentic workflows increasingly handle decision-making, task delegation, and execution, the boundary between computational architecture and social organization has blurred. In this context, re-examining the foundational principles of normative multi-agent systems offers a strong basis to identify the challenges for properly governing AI, and more generally socio-technical ecosystems.

Drawing from decades-long work in agent-based modeling within the legal domain, this chapter examines a number of relevant cross-cutting concerns, conceptualizations, modeling problems, and design challenges that apply equally to large-scale distributed Artificial Intelligence and to the agent-based modeling and simulation of social-institutional structures. Our aim is twofold: on the one hand, to provide architectural guidelines for moving beyond a contingent, task-centered, narrow view of AI; on the other hand, to reflect on the deep continuity between institutional and computational domains. However, even if we will explore concerns shared by AI and computer science, and cognitive, social, and legal sciences, we also acknowledge that cross-pollinations between these domains are driven by different motivations. For analytical purposes, this distinction can be formulated as two complex-adaptive-system design problems that draw on overlapping knowledge resources:
\begin{itemize}
    \item the problem of modeling, simulating, and reasoning about \textit{complex socio-institutional arrangements} like those expressed in law, markets, and accounted for by social stakeholders (institutions, businesses,  people) in general;
    \item the problem of designing \textit{complex information system arrangements} based on design metaphors drawing from vocabulary on minds and institutions.
\end{itemize}
Application domains like electronic institutions were particularly fascinating to researchers from both traditions because they required addressing both design problems simultaneously.\footnote{This is also why the chapter belongs to the present volume. It is not by chance that the notion of \textit{anchoring} presented in chapter 7 of this volume \citep{Noriega2026}---that is, the set of pragmatic
couplings that must hold for an online institution to be effective within the richer socio-economic-legal environment in which it is situated---matches with the notion of \textit{ecological validity} we will pursue in our contribution.} Today, emerging practices of digital governance are increasingly articulated through the language of \textit{agentic ecosystems}, \textit{governance harnesses}, and \textit{programmatic guardrails}.
These two once-parallel paths have essentially intersected: when an enterprise deploys an autonomous AI agent to execute financial contracts, perform medical triage, or manage supply chains, the underlying computational architecture enacts an institutional arrangement, even where that arrangement remains only partially aligned with the relevant legal, organizational, and social framework.

Our present aim therefore is to introduce the main challenges at stake on both sides in a way that inspires the reader to further reflection. For this purpose, we will prefer the construction of a high-level, conceptual framework and scope over an exhaustive literature review. Our central claim is that robust large-scale AI cannot be understood only as---in terms of today’s core Agentic AI components---model capability plus workflow orchestration. Where a distributed system acts in socially consequential settings, its architecture also allocates epistemic access, practical powers, duties, evidential burdens, and opportunities to contest decisions and revise procedures. These are institutional functions. Conversely, institutions increasingly need computational methods to represent social roles, diagnose failures, and test or revise procedures under varying conditions of complexity and pluralism.

The paper proceeds as follows. Section 2 will provide a historical reflection on the trajectory of the AI discipline, starting from understanding and modelling the ``mind'', suffering from the scaling problem, and focusing on engineering intelligence piece by piece. Section 3 will cover relevant perspectives on individual agency, while section 4 will focus on institutions and collective agency. Section 5 examines several forms of pluralism and connects them to ecological diversity. In the concluding discussion section (Section 6), we identify a number of potential grounds for novel advances in the application of institutional metaphors to distributed AI, and the application of AI techniques in modeling, analyzing, and evaluating socio-institutional arrangements.

\section{From modelling to engineering, from individuals to ecologies}
\subsection{Modeling the Mind} As a discipline, Artificial Intelligence (AI) started out with the ambition to reproduce the human mind as a monolithic computer program. The problem of reproducing intelligence in technology was approached primarily as a \textit{functional decomposition} exercise, with ``intelligence'' serving as the high-level function to be recreated within a software system by combining smaller functions realized through simple, well-understood input-output modules. It was only later that machine learning approaches rose to dominance, eventually becoming nearly synonymous with AI today; these focus primarily on the \textit{functional reproduction} of intelligent behavior, rather than its structured, controlled decomposition.

The nature of constituent primitive modules was rather clear to early computer scientists: \textit{logical inference} and \textit{knowledge structures} were expected to play a key role, informed by introspection, common-sense psychology, and contemporary concepts of rationality \citep{Newell1976}. The \textit{General Problem Solver} \citep{Newell1959a} may be considered the archetype of this approach and is frequently cited as such in lectures on the history of AI. In architectures of this type, scaling up from simple toy examples of reasoning to plausible intelligent function, even within restricted cognitive niches, becomes a challenge of managing complex knowledge in a logic-based representation language, where the formal properties of the logic enforce modularity and prevent unwanted interactions between represented knowledge structures. 

As soon as this required modularity breaks down---that is, when knowledge structures interact in unforeseen and undesirable ways---the construction of the higher-level function (intelligence) may catastrophically fail. If we are lucky, it fails obviously, such as when constraints enforced by the representation language are violated; more insidiously, it fails through unintelligent behavior in many other cases. From an engineering perspective, an AI system following this architectural paradigm lacks sufficient adaptive capacity, or resilience, to reliably incorporate new knowledge structures.

\subsection{The Scaling Problem, and Beyond It}
The scaling problem in an architecture based on logical inference and knowledge structures, continually presents itself as a \textit{knowledge representation} bottleneck: background knowledge is always too sparse to properly scope knowledge structures, causing the system to eventually draw inferences that are untrue and, more critically, obviously unintelligent. Historically, this limitation led to several distinct tracks of AI research:

\begin{itemize}
    \item to uncover and formalize implicit background assumptions---such as the infamous \textit{frame problem} \citep{Morgenstern1996}---that are typically overlooked and threaten system modularity;
    \item to codify massive volumes of \textit{common-sense background knowledge} for general AI use \citep{Lenat1995}, in the hope of eventually achieving critical mass;
    \item to identify \textit{foundational knowledge structures} with universal applicability---codifying them into formal top-level ontologies \citep{Sowa1995} for reuse across distributed modules; and
    \item to scope \textit{defeasibility} in logical reasoning \citep{Pollock1975} by designing non-monotonic logics and representation languages that contain undesirable interactions between knowledge structures.
\end{itemize}

Research that addresses these problems belongs historically to the \emph{symbolic} AI tradition. Perhaps its main impact was to steer the mainstream philosophy of the field away from the initial strict mathematical logic---and a disdain for logical inconsistency and irrationality---toward more psychologically valid views of reasoning and human rationality.

A further, partly complementary response to the scaling problem was to decompose the AI problem into two more manageable challenges, allowing AI researchers to focus on either:
\begin{enumerate}
    \item studying well-defined, \textit{specialized intelligent behaviors} (e.g., playing chess, recognizing human faces, or solving job-shop scheduling problems) within specific contexts of use; or
    \item proposing overarching \textit{architectural concepts} for interfacing specialized intelligent components. These are conceived of as agents---each competent in a limited set of behaviors---operating within a virtual organization to project the appearance of a single, generally intelligent agent to the outside world.
\end{enumerate}
We examine the impact of these two research trajectories in more detail.

\subsubsection{On specialized intelligent behaviours}
Solving the first type of problem yields direct, viable commercial applications that justify research funding, though in the past it was often dismissed as ``not real AI'' due to its lack of adaptive potential. Yet, ironically, contemporary public usage of the term AI overwhelmingly refers to applications of this kind.  Chatbots based on Large Language Models (LLMs), for example,
are typically centred on a large neural model trained to predict and generate text, embedded within architectural layers that sustain dialogue,
manage interaction, and shape outputs through instruction tuning and alignment methods such as \textit{reinforcement learning from human
feedback} (RLHF) \citep{Ouyang2022}. Nothing in these specialized systems explicitly addresses general intelligence. What appears to be general intelligence is, in fact, a side-effect of language being a general-purpose human interface. Yet, the possibility of codifying massive background knowledge brought the consequence that the search for universally appealing knowledge structures has migrated from top ontologies or explicit common-sense knowledge representations (a core research track in symbolic AI) to learned latent representations (although at the price of opacity, hallucinations, and possibly ill-formed structures).

\subsubsection{On organizational architectures}

Addressing the second challenge---proposing overarching architectures for agent coordination---historically received far less attention than the first one beyond AI research. These architectural questions gained public visibility through debates on the \emph{responsibility gap} created by the widespread deployment of autonomous and distributed systems, such as algorithmic trading agents, autonomous vehicles, and weapons systems like drone swarms \citep{Matthias2004}. They have now re-emerged as a practical, though still
largely under-theorised, concern in the development of modern \emph{Agentic AI} systems. 

Broadly defined, ``Agentic AI'' refers to AI systems engineered to operate autonomously or semi-autonomously (with minimal user intervention). Moving beyond passive, turn-based text generation, these systems proactively plan, reason, and direct multi-step workflows while adapting to dynamic environments to achieve higher-level goals. As this field evolves, new terminology is rapidly emerging---often shaped through bottom-up engineering practices rather than structured theoretical reflection. Central to this landscape is the concept of \textit{agent harness} (or \textit{scaffolding}): the software architecture that encapsulates core foundation models---most importantly \textit{large language models} (LLMs) or multimodal models with LLM heritage. In essence, a harness acts as the execution engine, managing state and memory, driving the decision-making control loop, executing tool calls, handling error recovery, and enforcing security boundaries. 

From this point of view, modern agentic architectures synthesize both historical strains of AI research: the underlying foundation models handle \textit{functional reproduction}, whereas the agent harness provides an operational \textit{functional decomposition} framework. Reframing these developments through debates on multi-agent coordination and normative systems can therefore provide stronger conceptual foundations
for their design and governance.

\subsection{Engineering a Mind as an Ecology of Minds}

For the purposes of this chapter, we are mainly concerned with the second challenge: overarching architectural concepts. Archetypal for this response are the agents in the visionary \emph{Society of Mind} of \citet{Minsky1986} (``{The power of intelligence stems from our vast diversity, not from any single, perfect principle}'', p. 308), as well as the concrete \emph{intelligent creatures} proposed by \citet{Brooks1991} in his seminal work ``Intelligence without representation''. According to Brooks, the fundamental decomposition of intelligence should not rely on modules that interface via a shared, centralized knowledge representation; rather, it should consist of independent \emph{creatures} that interact directly with their environment through perception and action, exhibiting ecological validity within that domain. The central insight here is to bring to the foreground the tight coupling between an agent---or creature---and its host environment, niche, or ecology.

Specialized intelligent behaviors within a specific niche can often be successfully isolated. Some problems, such as playing chess, may be tackled using specialized, explicit knowledge representations paired with appropriate search algorithms. Others, such as face recognition, are less amenable to solution through introspection; in AI systems, they are achieved not through knowledge representation techniques, but via machine learning algorithms operating on vast datasets of labeled examples that exemplify the behavior to be acquired. Large Language Models (LLMs)---operating fundamentally as systems optimized for text completion---fall directly into this category. However, there is a key difference: cracking a problem like chess as a symbolic search problem means becoming able to generate at need ecologically valid labeled examples---essentially \textit{trusted synthetic data}. This data can be subsequently used for machine learning, resulting in efficient, scalable machine learning chess with comparable performance. The reverse direction remains very difficult: recovering an explicit and accountable symbolic representation from an
opaque learned model denotes an unresolved explainability problem, addressed in the neuro-symbolic research stream. The third route---using the outputs of one learned model as unexamined training data
for another, is problematic. Recursive reliance on synthetic
data may produce unintentional and unrecognised distributional distortions
or \textit{model collapse} \citep{Shumailov2024}; where the generated data is maliciously
manipulated or insufficiently controlled, it may also create
\textit{data-poisoning} risks.  This strengthens the case for accountable generative mechanisms of behaviour where they can be constructed.



\subsubsection{Multi-Agent Systems, and IBM's Watson, and beyond}
Historically, the challenge of finding architectural concepts to create and maintain the appearance of a single, unified agent led to the development of \textit{multi-agent systems} (MAS). These architectures consist of multiple autonomous agents interacting within a shared environment, each operating with its own perception–action loop and without global knowledge of the system as a whole  (e.g. \cite{ferber1999}). A central concern in decentralised MAS research
is to avoid reliance on a single controller or globally shared memory,
because such arrangements may recreate the brittleness associated with
centralised functional decomposition. Researchers investigated decentralized organizational models and peer-level coordination mechanisms, aimed at enhancing overall system resilience and adaptive capacity (e.g. \cite{wooldridge2009}). Yet, although multi-agent system research has yielded many valuable theoretical insights into organizational dynamics (e.g. \cite{grossi2006,dignum2009}), few examples of practical value successfully combined relevant sets of intelligent behaviors within this traditional paradigm. It is instructive, however, to examine the first well-known system that demonstrated significant success in matching specialized computational creatures to specific problem niches: IBM's Watson \citep{Ferrucci2010}. 

Watson gained widespread fame for its decisive victory on the game show \emph{Jeopardy!}. From our architectural perspective, Watson performed essentially one core task: question answering (QA). Yet, its internal architecture operated as a coalition of distinct QA agents that \emph{compete} against one another to generate the highest-scoring response. In its essence, the system relied on giving conditionally attention to different representation subspaces embodied by the different `creatures'.\footnote{Strictly speaking, Watson was not a decentralised multi-agent system in the strong sense. Yet, it still illustrates how specialised computational perspectives can be comparatively evaluated rather than collapsed into a single inference chain.} 
Through the competitive dynamic, the system evaluated the candidate responses and learned to select the optimal answer based on features extracted from both the query and the candidate solutions. This competitive architecture represented a compelling departure from brittle functional decomposition, pointing instead toward more robust organizational metaphors. Related principles of specialisation, conditional selection, or
comparative evaluation appear across machine-learning paradigms, including \textit{generative adversarial networks} (GANs) \citep{Goodfellow2014},  ensemble techniques like \textit{random decision forests} \citep{Ho1995}, \textit{mixture of experts} (MoEs) \citep{Jacobs1991,Shazeer2017}, \textit{actor-critic reinforcement learning} \citep{Konda1999}. However, the structural and social scaffolding governing these statistical models remain comparatively minimal. 

In some respects, Watson's ensemble of specialised question-answering components anticipates patterns now found in LLM-based agent orchestration. While Watson relied on hand-crafted feature extractors and a statistically weighted arbiter within a closed domain, contemporary Agentic AI frameworks rely on dynamic, contingently instantiated (and dissolved) ``societies'' or ensembles of LLM-powered components, in which distinct personas or specialized sub-agents debate, critique, and refine intermediate outputs through an overarching \textit{agent harness}. These components
are often described as agents, even when their autonomy remains limited
and the overall workflow is centrally orchestrated.
Multi-head attention allows a LLM to attend jointly to different representational aspects of its input; multi-agent orchestration of LLM-powered agents can instead distribute work across differentiated prompts, tools, models, or perspectives. 

In short, by coupling decentralized multi-agent coordination with state/memory management and external tool access, contemporary Agentic AI implicitly attempts to realize the organizational metaphors envisioned by multi-agent systems research, transforming static statistical text generation into adaptive, goal-directed problem solving. It is therefore crucial to bring this vision to the foreground---both to re-examine the identified challenges and to chart a principled way forward.

\subsubsection{Agent-Based Modelling, and beyond}

Multi-agent system technologies and concepts have thus far achieved their greatest success indirectly, serving as the foundation for the \emph{agent-based modeling} (ABM) paradigm, with notable applications in understanding, simulating, and, in controlled settings, anticipating the macro-behavior of real-world socio-legal structures \citep{Railsback2019,conte2014agent}. However, agent-based modeling addresses a fundamentally different class of problems---in many ways dual to the ecological perspective outlined above---concerning the internal primitives required to specify an individual agent's behavior. Whereas specifications based on simple reflex architectures or quantitative utility functions may suffice for studying aggregate systemic phenomena under
stylised assumptions of economic rationality, human discourse concerning social actors typically invokes intentional and normative categories, prioritizing qualitative dimensions of behavior. 

To bridge this gap, research since the early 1990s has advanced the study of \emph{cognitive agents} (e.g. \cite{rao1992}) and \emph{normative multi-agent systems} (NMAS) (e.g. \cite{boella2006}), aiming to establish computational infrastructures grounded in inferential interpretive templates  drawn from practical reasoning, folk psychology, law, institutional economics, and linguistics.
The theory of mind tradition is exemplified by the \emph{belief-desire-intention} (BDI) agent architecture \citep{Bratman1987}, while formal frameworks for normative positions draw extensively on classical literature concerning norms \citep{Alchourron1971}, jural relations \citep{Hohfeld1913a}, and \textit{speech acts} \citep{Austin1975}. Crucially, the purpose of these formalisms is not necessarily to provide
empirically faithful models of human cognitive processing. Their purpose is to provide high-level and precise specifications of agent behaviour in
terms that remain intelligible to human participants, and that can guide or govern autonomous execution within complex socio-technical
arrangements. They aim, in other words, to make the relevant structures of agency explicit, inspectable, and open to revision.

\section{Agency as Unity}
\subsection{External and Internal Views}

In abstract terms, all problems reviewed in the previous section are expressions of a fundamental tension between \textit{unity} and \textit{multiplicity}. Multiplicity arises where unity fails, or to achieve what cannot be accomplished by individuals alone. Multiplicity is composed of unities which, when viewed in aggregate, may constitute a new unity; conversely, unities may recursively contain internal multiplicities. This ontological duality gives rise to two co-existing perspectives: the \textit{external} view (addressing a system-unity acting within an environment) and the \textit{internal} view (addressing a system composed of multiple interacting components). For example, an organization can be viewed externally as a single socio-economic actor that consumes and produces resources (e.g., products, services) to achieve specific goals; internally, it appears as an arrangement of roles embodied by individual employees, supported by a network of business units, subsidiary organizations, and coordinated partners. Similarly, a decision-support system or the guidance system of an autonomous vehicle may be perceived externally as a unified software artifact producing task-relevant outputs from given computational and informational resources, while internally functioning as a complex of interconnected modules providing the necessary execution functions. Unsurprisingly, this distinction between internal and external views is embedded in standard modeling practices (e.g., UML) and software engineering paradigms, reflected in concepts such as \textit{orchestration} and \textit{choreography}.

Yet this raises a fundamental question: Why is an entity perceived as a unity externally and as a multiplicity internally? More precisely, as we are dealing with entities designed to act within the world, what does it mean for a multi-component system to successfully project the appearance of a single agent to the outside world?
The question is not merely conceptual. In terms of legal accountability, for instance, obligations are ascribed to providers and deployers of an AI system as legally singular entities, whatever the internal multiplicity of the artefact. In terms of  operations, an agent harness may hide multi-component workflows behind a single interface; the system is thereby presented as one coherent agent, even where its internal operation is distributed.

The remainder of this section develops this distinction in two directions: first, by examining the interpretive conditions under which a complex
arrangement is treated as a unified agent; second, by identifying the agentive positions required to specify, explain, and govern its conduct.

\subsection{What Agents Look Like, Externally}\label{subsec:what-agents-look-like-externally}

When interpreting a scene, observers frame events depending on the most pertinent analytical perspective available to them. In a classic illustration by \citet{Hart1985}, a farmer may view a drought as the primary cause of a severe famine, whereas an international body might attribute responsibility to the host government for failing to establish adequate food reserves. Plausible criteria identified within cognitive science as governing an observer's choice of analytical stance include \textit{informativity}---the capacity to draw actionable conclusions tied to the observer's values or interests---and the required \textit{cognitive effort} (cf. \textit{relevance theory} \citep{Sperber1986} and \textit{simplicity theory} \citep{Dessalles2013}).

Among the interpretive attitudes available to an observer, Dennett's \textit{intentional stance} \citep{Dennett1987} is inherently more vulnerable to prediction errors than physical or design-based stances, yet it yields highly informative explanations at a fraction of the cognitive cost. For instance, explaining a child's movement toward a bowl of sweets by analyzing their underlying physiological state or evolutionary history would ordinarily be far less
tractable, and far less informative for practical purposes, than
attributing a desire for the sweets and a belief that they are available. The same economy of explanation often governs the external interpretation of complex collectives---such as organisations and nation-states---as well as the interpretation of AI systems as unified intentional agents.

\subsection{Intentional Ascriptions and Articulations as Reasons}\label{sec:articulations}

In adopting an intentional stance, observers attribute beliefs, desires, and intentions to an agent, treating it as a \emph{rational} entity---namely, one that ``will act to further its goals in the light of its beliefs'' \citep{Dennett1987}---to construct a reasonable approximation of its trajectory. Viewing intentional ascription as the attribution of \emph{reasons} for
conduct connects it to the broader phenomenon of \emph{mentalisation}
\citep{Fonagy1997}: interpreting behaviour in terms of beliefs, desires,
intentions, and other purposive states. In formal settings, the same move
supports a \emph{rationalisation} of behaviour, namely the reconstruction
of the reasons by which conduct can be explained or justified. Ultimately, an entity---whether a person, an institution, or an artificial system---is interpreted as an agent because observers can successfully ascribe beliefs, desires, and intentions to it in order to yield reliable predictions. The same dynamic holds introspectively when an agent observes its own behavior, articulating its internal motivations as explicit reasons for action.

Both \emph{having} articulated beliefs, desires, and intentions, and being \emph{able} to ascribe beliefs, desires, and intentions to another agent clearly facilitate communication among social entities. \citet{Mercier2011} make this point forcefully: evolutionarily speaking, the primary function of reasoning using explicit knowledge structures \emph{must be} argumentation. This insight is supported both by the historical emergence of human socio-institutional systems and by the increasingly central role that multi-agent deliberation plays in contemporary Agentic AI.

However, a different outcome might be expected depending on whether we adopt the agent or the observer point of view. Traditional experiments on \emph{confirmation bias} in human reasoning consistently reveal a striking asymmetry between producing and evaluating arguments: when generating arguments \emph{for} a claim, individuals exhibit a strong confirmation bias, selectively gathering and remembering supporting evidence; conversely, when evaluating or arguing \emph{against} a claim, they become remarkably adept at detecting and refuting weak, biased arguments. \citet{Mercier2011} connect these psychological findings to the rationality of dialectical argumentation---such as in a \textit{legal trial}---where a process governed by adversarial competition systematically promotes discovering truth, even when individual participants are merely striving to win.  

A central mechanism of any court (or dialectal) setting is the \emph{burden of proof}---the procedural obligation to supply evidence for asserted claims \citep{Walton1988}. Burdens of proof are systematically allocated to agents based on whether establishing a claim advances an actor's dialectical position. Within legal norm structures, these obligations can generally be allocated statically to the propositional conditions of a rule. When viewed through the lens of the burden of proof, the charge of individual cognitive irrationality in argument production loses its force: generating biased, self-interested arguments is not a flaw, but a functional strategy within an adversarial protocol.\footnote{In this light, behaviours routinely diagnosed as problematic in LLMs,
such as post-hoc rationalisation or the production of plausible but weakly grounded reasons, appear not to be isolated defects of
generation, but resulting from the alignment with the role of a producer of arguments.} 

The central question of ``rationalization'' shifts therefore to the design of the communication setting, the allocation of burdens, and the identification of those charged with arguing against a claim \citep{gordon2007carneades}. Institutionally, this concerns how normative rules can be structurally attached to particular communication settings, and which elements of a rule serve which institutional interests. 
To preserve a separation between argument production and evaluation, evaluators should not have access to precisely the same reasons available to producers, and vice versa. 

In an Agentic AI system implementing a 
\textit{generate \& verify} orchestration, as long as the verifier is looking at the same buffer and employs the same LLM (so it shares parametric knowledge) of the generator, it will take all input as its own and be defensive about it. For an argumentative setting to function, the verifier must maintain some form of \textit{information asymmetry}---having access to only a limited portion of the generator's knowledge.\footnote{See for instance \citep{xu2025}, showing that candidates who use the same LLM to write or tailor their resume as the one used by the recruiter/evaluator experience a 23\% to 60\% increase in shortlist likelihood, regardless of underlying differences in human qualification. } In other words, effectively challenging arguments requires some degree of independence.

\subsection{Scripting Behaviour and the Resulting Ecology}
But how can an observer ascribe a determinate course of action to an agent when the space of behaviour open to that agent is, for practical purposes, vast and underdetermined? In traditional decision theory, a rational decision-maker selects a solution by maximizing expected utility over candidate options. This optimization principle underlies the \emph{homo oeconomicus} view in classical economics: actors are assumed to be self-interested entities that systematically maximize their utility functions. 

There are many ways in which one can frame criticisms against these assumptions and account instead for the boundedness of human rationality, starting from the obvious observation that decision problems in the real world are often too complex, making it practically impossible to perform maximization over all potential alternatives. Heuristics, rules, and procedures emerge as an agent's effective responses to unpredictable errors in selecting optimal solutions. \citet{Heiner1983} identifies the root of this general phenomenon in the gap between an agent's competence at solving a problem and the difficulty of that problem within a given environment---the \emph{competence-difficulty} (C-D) gap.

Heiner convincingly argues that under large C-D gaps, an agent generally performs more effectively and efficiently over time by following simple rules rather than attempting full optimization. For instance, a skilled chess player tends to play routine openings to capitalize on past experience with the board states that develop from those lines, exercising greater creativity only in the middle and late game---even though standard interpretations of rationality might suggest that calculating optimal moves becomes easier as board complexity decreases late in the game. In social settings, this dynamic leads to the institutionalization of interactions, as participants perform better when behavior remains mutual and predictable. Similarly, while there are numerous valid ways to conclude a commercial transaction, a supermarket accommodates only a single routine to structure the interaction and minimize cognitive load. The result is that the agent engages in a form of \emph{ecological niche construction} \citep{Bardone2011}, associating relatively simple rule sets with the social roles it enacts when interacting with an environment, depending on its model of that environment.

\subsection{Social Intelligence as Compressing Behaviour}

In simplified terms, an agent and its environment can be modeled as interacting programs that exchange discrete symbols, where the environment emits reward signals in response to the agent's actions. Along related lines, \citet{Legg2007} define intelligent behavior as an agent's expected reward within a given environment, extending this to universal intelligence when evaluated across all computable environments. The agent's core task is to predict which output sequences will maximize cumulative future reward---a formulation that encompasses domains such as strategic game-playing (where competence relies on inferring an opponent's strategy), as well as question-answering (predicting responses that optimize evaluative feedback), text completion, and, at higher level of complexity, maintaining its functions in an open environment. 

To predict competently, the agent must effectively infer the underlying generative program executing within its environment. This challenge is formally equivalent to data compression: discovering a minimal program (the predictive model) capable of reproducing the observed environmental dynamics. This equivalence motivated the creation of the Hutter Prize \citep{Hutter2006}, which rewards advances in compressing large text corpora under the premise that compression efficiency is intrinsically tied to general intelligence. A similar intuition is formalized within \emph{algorithmic information theory}, captured by Chaitin's dictum that ``understanding is compression'' \citep{Chaitin2005}. Today, the ``intelligence as compression'' has transitioned from theoretical algorithmic information theory to a central tenet of modern technology: LLMs are, at their core, massive lossy compression engines that compress internet-scale human text into neural network parameters to predict subsequent tokens. Yet, despite being a powerful abstraction, \emph{intelligence as compression} has clear limitations from an ecological point of view \citep{Dowe2011}.\footnote{Already at an architectural level, the scale of data against which LLMs are constructed by brute force is the opposite of what one would expect from learning within ecological niches.} 

First, compression alone does not determine what should count as relevant signal, valuable performance, or acceptable error, i.e. reward mechanisms. In lossy compression schemes (such as audio and video compression), we discard parts of environmental data as noise rather than signal because they are irrelevant to the human observer---that is, no reward penalty is incurred. Contemporary LLM-centered systems explicitly address this by incorporating internal or external alignment mechanisms (such as \textit{RLHF}, \textit{direct preference optimizations}, or ``constitutional'' guardrails) to supply this evaluative signal. Yet, without an inherent external frame of reference, we have no theoretical account of how an autonomous agent distinguishes between useful signal and irrelevant noise; attempting to compress everything universally is ecologically invalid as a response to the cognitive limitations of an agent operating within a complex environment. In short, a comprehensive theory of intelligence must directly address Heiner's C-D gap.

The second limitation deals with social environments, and is in this context obvious. The intelligence-as-compression concept accounts for learning by induction from examples, but not for social learning through interaction with other agents. It cannot account for shared vocabulary between agents (i.e., an \textit{ontology}), nor with argumentation, and not with the exchange of instructions and normative rules between agents. Any account of these omissions would need to explain how agents, despite their diverse experiences, keep their programs sufficiently similar to exchange pieces of them with other agents.

In contemporary AI, this problem cannot be resolved by internal model alignment alone, as it intrinsically concerns the external, inter-subjective manifestations of agency. Nor is it covered by ``constitutional AI'' approaches, as these techniques are conceived for an individual model to critique and refine its own outputs, rather than to orchestrate a heterogeneous collective of social actors. Current attempts to supplement frozen foundation models with memory, interaction, or multi-agent feedback remain limited in their capacity for reciprocal and institutionally structured adaptation.
Designing an institutional architecture---including open communication protocols, contractual mechanisms, and normative governance frameworks---is one of the core concerns of \textit{normative multi-agent systems} (NMAS).

Nevertheless, the core insight of the intelligence-as-compression paradigm remains of interest even with NMAS: when confronted with a complex environment or a peer agent, intelligence entails the capacity to construct or select a generative program that accurately models that entity, alongside a correlative program (or policy) designed to drive fulfilling and rewarding interactions with it.

\subsection{Specifying Intentional Agents}
The traditional AI perspective internalizes the intentional stance and considers agents as intelligent systems---entities performing certain actions to achieve specific goals, conditioned upon their knowledge. Consider, for instance, the uncompromisingly external perspective towards agents taken by Newell in his seminal paper on the \emph{knowledge level} \citep{Newell1982}. Knowledge and goals are \emph{ascribed} to an agent to explain its behavior, remaining separable from the structures and mechanisms that create that behavior. In a sense, Newell proposes to create \emph{models} of observed agent behavior within the bounds of some design constraints, operating via a model of deliberation loosely referred to as the \emph{principle of rationality}. The question of how to reproduce the mechanisms that create the behavior is secondary, and in practice often easier if the design problem is well-defined and the goals of the agent are easy to identify (such as in constrained domains like chess or facial recognition).

In more open-ended settings, where goals are dynamic and need to be continuously re-set, this simplification no longer holds. Connecting goals to the broader category of desires, we can approach the traditional philosophical account of practical rationality based on a \textit{belief-desire} (BD) architecture. \citet{Bratman1987} convincingly argued against such two-parameter characterizations, highlighting the crucial role of \textit{intentions} (I). In a BDI framework, agents may hold conflicting, inconsistent desires---such as desiring both to eat chocolate and to adhere to a weight-loss plan---yet the subset of desires eventually selected to govern actual conduct must be mutually consistent against the backdrop of beliefs. It is this active selection, reified as intention, that captures the distinctively \textit{deliberative} aspect of agency.

However, for the reasons outlined above, in contrast to the traditional view of a deliberative agent---according to which agents make rational decisions by weighing competing reasons---one could instead start from an \emph{interpretative} basis, considering that agents reproduce established courses of action (or \emph{scripts}).\footnote{This role is played in contemporary Agentic AI by \textit{standard operating procedures} (SOPs) or workflow-centred specifications.} Scripts sediment deliberations that have  occurred previously: they associate recurrent situations with socially or
operationally established responses. Intentions, as commitments to specific courses of action, are then reduced once more to lower-order (concrete, contingent) volitional states selected by higher-order (generic, structural) desires in accordance with the abilities and knowledge ascribed to the agent, alongside the perceived state of the environment.

Evidently, insights gained from an interpretative standpoint can be remapped in generative terms: viewed from within the agent's internal architecture, these same mental states serve as the drivers of its conduct. An agent behaves in a specific manner precisely to \emph{satisfy} its intents. Whether such mental states possess literal ontological status or are merely epiphenomena of underlying neural or computational processes is irrelevant with respect to their functional utility. Indeed, the foundational principle of \emph{responsibility} in socio-legal systems relies on the pragmatic assumption that such intentional states are indispensable for reasoning about human behavior.

\subsection{Agentive Positions\label{sec:agentive-positions}}

Upon closer inspection, the standard \emph{belief-desire-intention} (BDI) triad can be further expanded. The classical BDI template lacks for instance explicit primitives for representing \emph{agentic abilities} (which condition the action-selection process) and \emph{sensory capabilities} (which determine which elements of the environment are perceptible to the agent).\footnote{Computational platforms for executing BDI agent scripts were obliged to introduce such mechanisms; see, for instance, \textit{goal-plan} and \textit{belief-plan rules} in AgentSpeak(L)/Jason. Conversely, most of these platforms adopted a flattened view of desires, treating them exclusively as executable goals (see e.g. the discussion in \cite{mohajeriparizi2022preference}).} To bridge this gap, in previous work \citep{Sileno2015a,Sileno2016} we recognized four primitive categories required to specify the behavior of an agent: \emph{commitments}, \emph{affordances}, \emph{expectations},  and \emph{susceptibilities} (CAES). Informally, these correspond to what the agent wants, what (it perceives) it is able to do, what it believes, and what it is disposed to react to. Any instance of these classes constitutes a \emph{position} assumed by the agent with respect to its environment. Introducing these positions here is relevant here as they hold a direct counterpart with institutional positions. 

\textbf{Commitments} ($C$) encompass motivational states eventually driving action. These include commitments that are \emph{acceptable} to the agent (desires), those that are \emph{preferred}, and finally those that are \emph{selected} as objectives or \emph{intents}, which subsequently trigger an \textit{intention} associated with a specific course of action (or \textit{plan}). Technically, introducing prioritization at the preference level resolves potential internal inconsistencies, enabling a conflict-free selection of desires to be committed to as intents. From a generative perspective, commitments can be interpreted as agentic dispositions possessing sufficient stability to drive sustained conduct. 
Moreover, reasons for action do not always reduce to concrete objectives; desires promoting abstract \emph{values} profoundly influence agent behavior, albeit operating at a deeper level of rationalization. \textbf{Expectations} ($E$) reflect the \emph{situatedness} of the subject in the world, capturing what the agent believes the world is, both \emph{actually} and \emph{potentially}. The actual category mirrors traditional beliefs, whereas the potential category designates what the agent projects the world may bring about under specific conditions, incorporating causal dynamics. Within the intelligence-as-compression hypothesis, expectations fulfill the role of the generative program executed by the environment, which the agent leverages to maximize rewards relative to its active commitments. \textbf{Affordances} (or abilities, $A$) represent action opportunities---the possibilities for an agent to enact specific behaviors under given environmental conditions to achieve designated outcomes. Affordances interact dynamically with commitments to determine which behavioral trajectory is ultimately selected. \textbf{Susceptibilities} (or sensory capabilities, $S$) capture the \emph{reactive} dimension of agency; an agent exhibits susceptibility to a given event if it registers a reaction to its occurrence at either the epistemic or motivational level.

The CAES template supports both explanatory and generative inferences. At
its core lies the possibility of a conflict between an active commitment
and an expectation concerning the current state of the world. From an
explanatory perspective, such a conflict can support \textit{abductive
intent ascription}: it helps explain why an agent acted. From a generative
perspective, it can initiate \textit{action selection}: the agent searches
for an available course of action capable of reducing or resolving the
conflict. Beyond this foundational control loop, the CAES template enables us to capture additional functional dependencies, including:
\begin{itemize}
    \item \emph{enactive perception} (mapping from active commitments $C$ to a susceptibility $S$): perceptual experience must be coupled with the agent's active course of action, primarily to anticipate potential (often negative) outcomes through continuous monitoring;
    \item \emph{affordance activation} (mapping from active expectations $E$ to an affordance $A$): because epistemic conflicts are evaluated at a higher level, an auxiliary informational channel is required to contextualize the agent's capabilities within the current situation;
    \item \emph{affordance inhibition} (mapping from a stronger commitment $C'$ to $A$): when a contemplated action generates consequences that conflict with higher-priority desires, the associated affordance is inhibited to prevent execution, even if perceived.
\end{itemize}
These dependencies can be related, at a high level of abstraction, to techniques found in contemporary Agentic AI. Enactive perception has loose analogues in relevance filtering, retrieval selection, and context-window management; affordance activation resembles context-sensitive tool availability, such as dynamic API-schema injection or state-dependent function calling; and affordance inhibition resembles programmatic safety
boundaries, including permission checks, execution interceptors, and guardrail mechanisms. The CAES framework is not reducible to these engineering techniques, however, because it specifies the functional roles that such mechanisms may play within a broader architecture of agency.

Consider, for instance, an assistant managing a task board. Initially, the agent possesses a basic affordance ($A$): posting tasks. When a user requests coding assistance, perceiving this input ($S$) leads the agent to commit to this request ($C$) and then to apply the associated affordance ($A$). Taking this commitment also produces explicit expectations ($E$) regarding task completion shared with the user, while dynamically activating new context-dependent affordances ($A$)---such as file system manipulation and local code execution. These tools remain available until an explicit discharge by the user ($S$) clears the task from the board, updating the agent's active commitments ($C$) and restoring baseline susceptibilities ($S$).

\section{Institutions and Collective Agency}\label{institutions-and-collective-agency}

\subsection{Going Internal: Control vs. Ecological Structures}

The external point of view serves as the basis for \textit{interpretation}, whereas the internal viewpoint is foundational for \emph{design}. In social and computational systems alike, the implementation layer consists of interacting actors, components, and
artifacts, some of which may possess varying degrees of autonomy (agents). System design therefore concerns establishing structural arrangements that enable and support those elements in forming and maintaining robust organizations to achieve higher-order goals. Such arrangements typically combine cooperation, coordination, and
competition, under conditions of bounded information exchange and only partial commitment to shared vocabularies and behavioural rules.

Evidently, distributing operational and computational loads across relatively
independent agents can shift system design away from tightly integrated
\emph{control} structures and toward more decentralised \emph{ecological}
structures. This fundamental shift is best illustrated by the philosophical distinction between \emph{totalities} and \emph{assemblages} proposed by \citet{Deleuze1980}, and further articulated by \citet{DeLanda2006}, hereby  summarized in the following table:

\begin{center}\vspace{7pt}
{\renewcommand{\arraystretch}{1.3}

\begin{tabular}{ll}
	\toprule
	\begin{minipage}[b]{0.46\columnwidth}\raggedright
		\emph{Totality}\strut
	\end{minipage} & \begin{minipage}[b]{0.46\columnwidth}\raggedright
	\emph{Assemblage}\strut
\end{minipage}\tabularnewline
\midrule
\begin{minipage}[t]{0.46\columnwidth}\raggedright
	organic composition (e.g.~heart in body)\strut
\end{minipage} & \begin{minipage}[t]{0.46\columnwidth}\raggedright
ecological coupling (e.g.~symbiosis)\strut
\end{minipage}\tabularnewline
\begin{minipage}[t]{0.46\columnwidth}\raggedright
	components specified by relations of \emph{interiority}: all their
	properties are manifest\strut
\end{minipage} & \begin{minipage}[t]{0.47\columnwidth}\raggedright
components specified by relations of \emph{exteriority}: only part of
their properties is manifest\strut
\end{minipage}\tabularnewline
\begin{minipage}[t]{0.45\columnwidth}\raggedright
	components derive their identity and function primarily from their place in
the system \strut
\end{minipage} & \begin{minipage}[t]{0.45\columnwidth}\raggedright
components retain capacities and dispositions that may be exercised in other assemblages\strut
\end{minipage}\tabularnewline
\begin{minipage}[t]{0.45\columnwidth}\raggedright
	dependencies logically necessary\strut
\end{minipage} & \begin{minipage}[t]{0.45\columnwidth}\raggedright
dependencies contextually obligatory\strut
\end{minipage}\tabularnewline
\begin{minipage}[t]{0.45\columnwidth}\raggedright
	failures compromise the system\strut
\end{minipage} & \begin{minipage}[t]{0.45\columnwidth}\raggedright
failures irritate the system\strut
\end{minipage}\tabularnewline
\bottomrule
\end{tabular}
}\vspace{7pt}
\end{center}

The passage from totality to assemblage requires several conceptual steps. For the sake of the argument, let us imagine that we start from a monolithic software application implementing an IT service, made up of several internal modules. The first step is introducing redundancy into the system, i.e., components potentially providing similar functions. If there is only one module that implements a required function for the IT service to function and this module fails, the application will stop working properly. If there are multiple modules that might be invoked to provide overlapping functions, the service resilience will generally increase and the problem will become rather one of an economic nature---i.e., of settling on an adequate resource distribution strategy among these modules. However, the IT service example still maintains a core module that is dispatching resources to the others, according to its requirements (just as MapReduce \citep{Dean2008} relies on a central scheduler for fault tolerance). As a second step, this constraint needs to be relaxed, by supposing an
assemblage without a core module: there is no central coordinator putting preferential requirements in the foreground. In nature, for instance, maintenance functions emerge as a \emph{selection effect}: assemblages that are unable to sustain themselves disappear.  In these conditions, components of an assemblage exist independently of whether the assemblage persists or dissolves. In other words, components of a totality are fully specified by their actual properties within the system (\emph{relations of interiority}), whereas components of an assemblage have dispositions which are not necessarily manifest in the context of that specific assemblage (\emph{relations of exteriority}).

Considering the components of the assemblage as agents, whose intent satisfaction counts as a reward obtained from the environment, we can outline the following general templates:

\begin{itemize}
\item
\emph{competition}: agents are committed to the same target, but the satisfaction of that target by one produces a physical or symbolic scarcity for the others;
\item
\emph{cooperation}: agents have dependent commitments, i.e., the satisfaction of one agent's intent is required for or facilitates the satisfaction of another agent's intent; symmetric dependencies form the basis of mutualism and enable maintenance functions (as in biological symbiosis);
\item
\emph{coordination}: a structured mechanism (resulting from individual or collective agency) distributes rewards and punishments to agents specifically to achieve higher-order goals.
\end{itemize}

A distinguishing factor of coordination schemes is the presence of explicit signaling. For instance, symbiosis persists even without explicit communication, whereas institutions cannot exist without some form of language (not necessarily verbal). 

\subsection{The Central Role of Failures} Institutions are indeed prototypical examples of mechanisms of coordination: they build upon symbolic means and form an infrastructure providing incentives (in the form of rewards and punishments, or the proverbial ``carrots'' and ``sticks'') to social actors, thereby modifying the incentives available from the non-institutional environment. Competitive dynamics are instead generally extra-institutional (e.g., the choice of a sale price in a market). 

The primary practical function of a legal system is not only to set expectations among social participants, but also to intervene when an institutional failure is presumed to occur in social interactions---i.e., when the institutional expectations of one of the parties involved are not met. \emph{Ex-post} judicial interpretations are meant to make explicit the normative positions of the parties prior to the failure, and then to associate an institutional response if the failure is confirmed, building upon the sources of law. Within this perspective, legal sources operate as reasons for enriching and
revising the behavioural models ascribed to the parties.

The centrality of failure is perhaps the strongest point of contrast between the institutional and the common engineering treatment of norms. Engineering practice---and machine-learning practice even more so---is organized around aggregate performance on the expected case.\footnote{Safety engineering and
incident-response practices are relevant exceptions.} A legal system is organized around the exceptional case, and specifically around the case in which someone complains. This is not a deficiency of either system; it reflects the fact that institutions exist to handle what could not be anticipated at design time, which is why the \textit{ex-post} interpretive moment cannot be compiled away. Any proposal to automate an institutional function must therefore answer a question that does not arise in ordinary system design: what happens when the automated determination is contested, and by what route does the contestation reach a party with the power to revise the determination?

\subsection{Normative positions through the Hohfeldian Prisms} Hohfeld's analysis of fundamental legal concepts \citep{Hohfeld1913a} starts from a similar interpretative consideration and distinguishes two families of legal relations holding between two social actors that are bound by legal provisions. The first concerns
\textit{obligative} positions---claims, duties, privileges or liberties, and
no-rights---which can be described as a \textit{deontic} dimension. The second
concerns positions of institutional change---powers, liabilities,
immunities, and disabilities---which can be described as a \textit{potestative} or
\textit{capacitative} dimension. This framework introduces two crucial innovations.

First, it posits that concrete normative relations cannot be expressed solely on the basis of deontic modalities. For operational purposes, it is crucial to make explicit which party is the \emph{addressee} of the normative proposition and which party is the \emph{beneficiary}---i.e., whose interests are protected or promoted (cf. the teleological aspect highlighted by \citet{Sartor2006}). Thus, the notion of a \emph{claim} embodies the idea of a right as the protection of an interest via a corresponding \emph{duty}. A \emph{privilege}, instead, corresponds to the absence of a duty, such that when it holds, the counterparty has \emph{no-claim} to advance. In previous work \citep{Sileno2014b}, noting the intuitive correspondence of duty with obligation and privilege with permission (in the colloquial sense of a \textit{faculty}, rather than its standard formal definition), we added a negative dimension to the traditional Hohfeldian square, obtaining the first Hohfeldian prism (Fig.~1, where A, E, and Y are the positions of the deontic triangle of contrariety).

\begin{figure}[t]
	\centering
	\scalebox{0.5}{
       \includegraphics{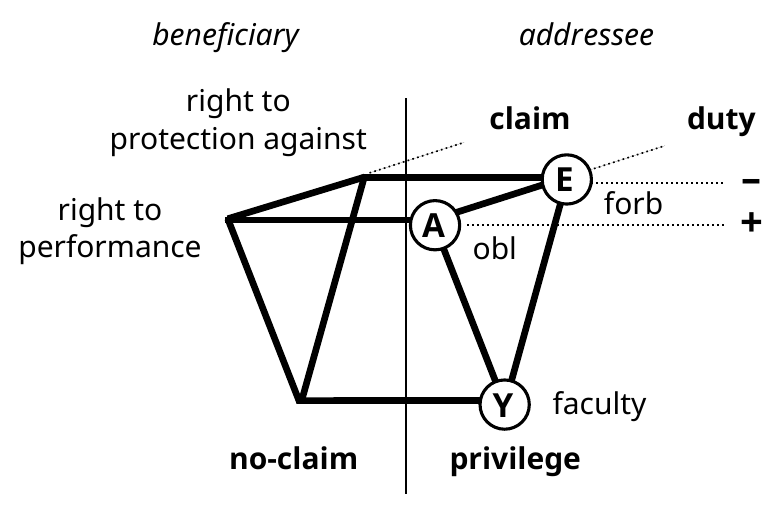}
	}
	\caption{First Hohfeldian prism, expressing deontic or obligative normative positions.}
\end{figure}

The second innovation is the explicit consideration of the dimension of institutional change, centered around the notion of \emph{power}. Hohfeld insists as well on the characterization of institutional power through volitional control, that is, through intentionality. A person holding such a power has the institutional \emph{ability} to deliberately alter legal relations (e.g., the transfer of ownership) by performing certain acts. Rather than using the terms \emph{addressee} and \emph{beneficiary} when considering the action, the two parties can be distinguished as the (potential) \emph{performer} (enacting the power) and \emph{recipient} (suffering the effects of the enactment). Correlative to power, \emph{liability} means being subjected to that power, while the opposite, \emph{immunity}, means to be kept institutionally untouched by the other party performing the action (who, in turn, holds a position of \emph{disability}). As before, by introducing a negative dimension we unveil the second Hohfeldian prism (Fig.~2), discovering the neglected positions of \emph{negative liability} and \emph{negative power}, which are relevant to undermining institutions. \footnote{For an extended analysis of the Hohfeldian prisms, see \citet[Ch.~4]{Sileno2016}.}

\begin{figure}[t]
	\centering
	\scalebox{0.5}{
		\includegraphics{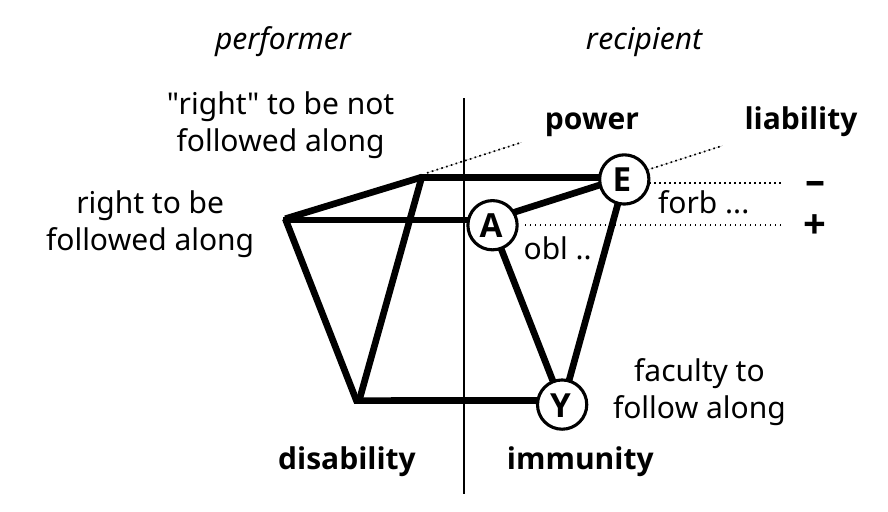}
	}
	\caption{Second Hohfeldian prism, expressing potestative normative positions.}
\end{figure}

The visual representation provided by the prisms (or squares, as in the original contribution) makes explicit the \emph{symmetry} and \emph{duality} of the relations between two parties.\footnote{A detailed  diagrammatic analysis has been given in \citep{pascucci2023}.} Focusing on a specific perspective (e.g., that of the addressee), two \emph{positions} (or three, if counting negative attitudes) are available to describe the situation in which that party stands with respect to another party. This position is intrinsically linked to the position of the other party. One may think of these diagrams as a game board where when a player moves, both checkers move simultaneously (\emph{correlativity axiom}). Thus, the difference between, for example, a duty and a claim is merely a matter of point of view, as both describe in principle the exact same binding. 

So-called ``objective'' rights illustrate however how the picture becomes more complex. Consider, for instance, property rights or rights to bodily
integrity, which may generate duties \textit{erga omnes}. The right-holder is easily identifiable, while the relevant duty-holders form an open and
potentially indeterminate class. In addition, the authority responsible for monitoring, enforcing, or remedying a violation may be distinct from
both the right-holder and the immediate duty-bearer. Closing this abstraction gap is necessary for the operationalisation of institutional
governance.

\subsection{Interface between Normative and Agentive Positions}

Interestingly, the four agentive categories of commitment, expectation, affordance, and susceptibility expressed in the CAES framework presented in section \ref{sec:agentive-positions} can be put in direct correspondence \citep{Sileno2015a} respectively with the four normative categories of duty, claim, power, and liability, or DCPL\footnote{These four categories are indeed at the base of the informational model captured with the DCPL language \citep{Sileno2022}, see \url{https://github.com/uva-cci/DCPLschema}.}, interpreting the environment as the correlative party to the agent.

This distinction can be traced as one of \emph{intrinsic} versus \emph{extrinsic} attitudes shaping an agent's behavior. Social norms (including legal norms) provide reasons for the agent---by attempting to influence the environmental feedback---to promote or demote certain action selections in its conduct (via obligations and prohibitions), or even to create the possibility of certain action selections (via institutional power).

Let us divide all institutional knowledge into class-norm ($N$) and token-fact ($F$) types of components. By modifying rewards provided by the environment, both norms and institutional facts supposedly play a role in the expectation category within an agent's reasoning, and consequently affecting the other categories. Intuitively, to facilitate normative alignment between agents, norms should be accessible to those expected to act under them, and therefore require a medium of transmission. The case of institutional facts is a bit different; to protect sensitive information, they should in principle be shared only on an institutional coordination basis. However, this distinction is not as rigid as it seems: the universal ambition of the modern rule of law can be seen rather as a special case---a consequence of its large base of addressees; companies do not necessarily want to share their policies widely, and social groups may have unique customs that are not explained to outsiders, with the effect of facilitating the distinction between in-group and out-of-group interactions.

Building on this framework, we can distinguish ten functional powers through which collective agency may transform commitments, expectations,
affordances, susceptibilities, norms, and institutional facts. The term
\emph{power} is used here in a broad functional sense; not every item is
itself a Hohfeldian legal power. Some concern the exercise of agency, whereas others concern the institutionalisation, publication,
interpretation, monitoring, or registration of its conditions.

\begin{itemize}
\item\emph{directive power}, consisting of setting a contingent commitment $C$;
\item\emph{operational power}, consisting of performing a course of action by means of the affordance $A$;
\item\emph{enabling power}, of activating the affordance $A$ (cf.~the dual \emph{disabling power});
\item\emph{regulatory power}, of transforming the commitment $C$ to the norm $N$ (in the sense of a structural commitment);
\item\emph{publishing power}, of reifying the norm $N$ on a publicly accessible artifact;
\item\emph{interpretative power}, of interpreting observations, producing beliefs and expectations $E$;
\item\emph{monitoring power}, of sensing a certain input by means of a susceptibility $S$;
\item\emph{attention power}, of activating the susceptibility $S$ (cf.~the dual \emph{diverting power});
\item\emph{declarative power}, of transforming the contingent belief $E$ into an institutional (structural) fact $F$;
\item\emph{registering power}, of reifying the fact $F$ as an artifact.
\end{itemize}  

These ten powers correspond to distinct administrative capabilities. For instance, in computational systems, directive and operational powers govern task execution and usage control; regulatory and publishing powers handle dynamic policy enactment and updates; and declarative and registering powers modify and manage shared distributed ledger states or databases. 

\subsection{Distributed Agency} Hypothesizing that these identified functions are correct, we should find them independently of whether the agency is implemented in a control-based or an ecological system, natural or artificial. In effect, one can recognize here a basis for the famous \emph{trias politica}. The \emph{executive}, \emph{legislative}, and \emph{judiciary} branches can be seen as concentrating, respectively, the directive and executive, the regulative and publishing, and the interpretative and declarative functions of a collective agency. The principle of the \emph{separation of powers} can be explained as aiming to maintain the balances present in the standard reasoning architecture. If the same actor holds, e.g., both the directive and interpretative powers, there is a short-circuit in the architecture because interpretation could be forced to enable this actor's commitments, rather than respecting those consolidated in the norms.  

Many interventions of the judiciary occur today on an asynchronous basis: the triggering event consists of a social actor (the plaintiff) that supposedly experienced an institutional failure and goes to the court. This means that monitoring power is left to individual social participants. However, the wide presence of global online platforms, as well as of institutional administrative digital repositories, has opened possibilities for upstream monitoring of several types of socio-economic transactions by entities having access to such information. \citet{DeMulder2012} have argued for an extension of the principle of the separation of powers to a \emph{tetras politica}, incorporating these potential monitoring branches. According to our architectural analysis, their argument is sound. 

In a distributed system, monitoring effort is supported by attention mechanisms. In human society, the actors that historically competed to hold attention power were media outlets. In contemporary ``attention'' economies, however, traditional media actors---which rely on human editorial curation and operate within established institutional and legal accountability frameworks---are increasingly superseded by algorithmically controlled media, typically centered on recommender systems and social media platforms. These digital media platforms modulate information flow in far more opaque and inaccessible ways;  although they are not wholly outside legal and regulatory frameworks, their forms of accountability remain partial, contested, and unevenly enforceable.

A further extension to our account involves focusing on the actors who maintain registering power. With the advent of regulatory technology (RegTech), distributed ledgers and smart contract infrastructures have increasingly entered the institutional operational cycle. Looking ahead, one can expect these to become relevant architectural abstractions for Agentic AI, necessary instruments to address and resolve conflicts in distributed transactions between autonomous agents.

\subsection{Enforcement as Maintenance Function} Although we have recognized functions in collective agencies that are ascribed to a single agent, we have not elaborated on how this is actually possible. For systemic maintenance, we expect the existence of some \emph{enforcement} mechanism (generally coupled with reinforcement) that is able to guide and maintain, to a large extent, individuals who are in principle independent to form an assemblage. Consider again the symbiosis example: the contextual obligation of interaction is a consequence of the positive reward of such an organization with respect to the absence of interaction.

Enforcement actions generalize this principle by either circumscribing undesirable behaviour through the prevention of its possibility (\emph{ex-ante} enforcement) or providing an institutional response to failures that strengthens general compliance through \emph{rewards} and \emph{sanctions} (\emph{ex-post} enforcement). Traditionally, \emph{ex-ante} enforcement serves as a core mechanism for authorization systems, whereas \emph{ex-post} enforcement appears more prominently in judicial systems.

However, enforcement is not necessarily driven solely by legal mechanisms, but also by social or physical dispositions (e.g., regulations regarding right-hand or left-hand traffic are backed by a form of ``natural'' enforcement: not complying with the correct driving direction carries direct and potentially severe consequences). Yet, legal transactions have the advantage of foregrounding explicit protections against the risks of a lack of coordination between the parties.

The general evolution of contemporary enforcement practices---observable not only in modern legal systems, but also in schools, parenting, and informal groupings---has been nicely synthesized by \citet{Dari-Mattiacci2013} as ``the rise of carrots and the decline of sticks''. The authors propose a theory of this process, showing that, from an economic perspective, punishment-based systems are more efficient in simple settings, i.e., when (i) the burden of compliance can be well specified (or, in other words, it is simple to verify whether a certain behavior is compliant with the rules) and (ii) is fairly distributed among the involved parties. Reward-based systems, on the other hand, are more efficient when the regulator has difficulties specifying the burden, or if the burden is highly skewed against a few social participants (becoming, in practice, a negative lottery). As modern societies become more complex, the heterogeneity of contexts and tasks is increasing, making reward-based systems increasingly prevalent.

But why does this happen? At first glance, we could write any contract either carrot-style or stick-style and they would be identical, at least from an algebraic point of view. However, experiments conducted in behavioral economics on framing have demonstrated that formally similar gains and losses are not experienced or
evaluated symmetrically \citep{Kahneman1979}.  
Even at the analytical level, Dari-Mattiacci and De Geest observe that there is an intrinsic difference due to the \emph{monitoring} component: if monitoring is not perfect, the two formulations are different because a party that is not monitored in a carrot-based system counts as non-compliant; whereas, in a stick-based system, it counts as compliant. In practice, as observed by \citet{Boer2014}, a reward regime assumes that most people are non-compliant; while a punishment regime assumes that most people are compliant. Boer then proposes the following \emph{evidence criteria} to decide whether an enforcement system is based on carrots or on sticks:
\begin{itemize}
    \item
    a \emph{reward regime} requires the production of evidence of
    compliance;
    \item
    a \emph{punishment regime} requires the production of evidence of
    non-compliance.
\end{itemize}
These criteria shift the attention from internal, subjective aspects relative to the parties to external aspects specified in regulations. Interestingly, these mechanisms are related to the \emph{burden of proof} (see section \ref{sec:articulations}), usually distributed among claimants (for evidence of non-compliance) and addressees (for evidence of compliance).

\section{Pluralism as Ecological Diversity}\label{pluralism-as-ecological-diversity}

\subsection{Epistemological Pluralism} Interestingly, the dichotomy totality versus assemblage also helps with understanding the distinction between formal and informal conceptual systems. The difficulty of constraining natural language in formal semantics is well known. The regularity of association of linguistic terms with a certain interpretation is a consequence of the relative regularity of the environment, of the sensory-motor capabilities, of the knowledge, and of the commitments of the communicating actors---but a change in one of those aspects might entail a change in those associations as well. Taking this phenomenon into account means accepting \emph{epistemological pluralism}.

Furthermore, beyond \emph{inter-agent} pluralism (agents, being independent, share neither the same knowledge nor the same interests or evaluative frameworks), there is also an \emph{intra-agent} pluralism to be considered, stemming from \emph{ontological stratification} \citep{Boer2009}. For instance, a road may be seen as a curved line while looking at directions on a map, as a surface while driving, or as a volume when laying asphalt to construct it. These alternative conceptual cuts, reminiscent of Dennett's stances, depend on which properties of the entity are relevant to the task at hand. Successful alignment of vocabularies or ontologies requires an adequate analysis of the underlying commitments associated with the task domain.

Normative systems, which are symbolically defined, inherently suffer from this problem, as the possible contexts in which they might operate cannot be fully anticipated. Yet at the same time, they provide a prototypical solution to the issue, by attempting to consolidate goals into a higher-order, abstract form, leaving further contextualization to be specified on demand.

\subsection{Normative Pluralism} But there is another level of assemblage to be taken into account for normativity: agents belong to a single normative system only in an ideal case. Think of all the rules and norms, written and unwritten, that we encounter in daily life. At home we perform certain chores because we at some point agreed to a distribution of them, or perhaps simply because they are expected of us. We followed both traffic rules and informal driving etiquette on the roads while traveling to and from work. We parked the car following rules and directions, greeted colleagues following social conventions, and perhaps followed directions about where not to smoke, where not to drink our coffee, and where to discard our paper coffee cup. Finally, we are writing this chapter following grammar and spelling rules, instructions about structuring chapters, and conventions about citation and plagiarism, all while working to meet a deadline. These norms belong to different types, and we recognize them as part of various complex normative systems. Some are based in law, others in the social conventions of society at large, in the conventions of the academic field, those of a particular building, or even of a small private circle. Any of these norms we may have noticed or overlooked, regarded or disregarded, and followed or violated, consciously or unconsciously. And for the most part we hardly think about them---a circumstance which, when we do pause to consider it, is truly remarkable.

Normative pluralism poses two central questions (see, e.g., \cite{Twining2010}). The first question is: \emph{how do we notice norms, and detect and resolve conflicts between pairs of norms as they arise}? When two norms are in conflict, they do so because they either:
\begin{itemize}
\item
disaffirm each other, i.e., appear to label the same intention as both strongly permitted and impermissible, or
\item
require us to adopt two intentions that are jointly inconsistent with our beliefs and thus cannot be executed simultaneously if our beliefs are correct---for example, requiring that we both keep our cake untouched and that we eat it.
\end{itemize}

Norms can be grouped into a \emph{normative order}---an institutionalized system of norms aimed at ordering social relationships among members of a community or group. Conflicts between these norms are generally ironed out within the system itself, with \textit{meta-rules} for determining the applicability and priority of norms. This is, however, only possible to the extent that agreed-upon norms for conflict resolution and prioritization can be established within the community or group, provided that recognizing norm applicability is not itself the overriding concern, and that the outcomes of the conflict resolution process remain acceptable to the members of the community (i.e., the right norms are applied).

\subsection{Value Pluralism} The second question posed by normative pluralism is: \emph{how do we measure how norms perform overall if they are to be effective in a plurality of norm environments?} Systems of norms range from public ones, with universal appeal, to the most private. Thus, the problem of measuring norm performance is relevant at all levels of granularity, and from both intra-agent and inter-agent perspectives. Individual agents, organizations, and whole communities form judgments about the performance of individual norms, and the problem of forming opinions about individual norms in a plurality of norm environments is rather ill-defined if it is the norm environment itself that frames what rational action is.

One might naively view this problem as purely quantitative---namely, as a matter of measuring overall levels of compliance. From the perspective of the values served by norms, however, this is typically ill-advised. A norm performs well if it functions---or is complied with---in the environments where it is useful, and fails or is violated in those environments where its value is outweighed by more important values served by competing norms that should take precedence. As such, aggregate compliance levels are meaningless unless properly scoped to the specific \textit{ecological niche} in which the norm ought to apply. 

Modern struggles in AI alignment illustrate this well. The EU AI Act states that ``AI systems shall be designed and developed in such a way that they achieve an appropriate level of accuracy'' (Art 15(1) of Regulation (EU) 2024/1689). This requirement rarely translates to a simple aggregate accuracy measure. The way of measuring, the scope of the measurements, and the threshold to clear are determined by niche, and some types of error (think of false positives vs. false negatives in simple yes/no decisions) clearly weigh far more heavily than others. 

Blanket enforcement of guardrails (against racism, uses for adult content creation, bomb-making, or hacking, etc.) on foundational models like LLMs without regard to uses in different niches is problematic. HuggingFace for instance recently made the news when it stated that it had to install a Chinese model (GLM) to be able to evade hacking guardrails in order to defend against being hacked by an OpenAI AI system\footnote{\url{https://huggingface.co/blog/security-incident-july-2026}}. The guardrails do not really contribute to compliance with the accuracy requirement, which applies to `AI systems' instead of models. Furthermore, multi-agent architectures typically increase the problem. If a model is directly interfaced to an end user a 90\% accuracy rate for a guardrail may seem superficially acceptable. But serially connect a speech-to-text, LLM, and text-to-speech model and the guardrails of the models may already falsely trip (assuming independence) up to approximately 30\% ($1 - 0.9^3$) of the time, and become a practical obstacle to third party development of agentic AI using commercially available models. 

\subsection{Pluralisms and Ecological Diversity} 
A general pattern can be traced at this point. Intra-agent epistemological pluralism emerges from simplifying the agent-environment coupling relative to functional niches, thereby driving the explicit articulation of multiple agent-environment couplings; in other words, a single agent may require different conceptualisations of the same environment for different tasks and roles. Inter-agent pluralism aggregates these articulations---and the commitments exploiting them---resulting in a set of \emph{agent-roles} available within a social niche. Normative and value pluralism can be seen as components of the former, aggregating \emph{social coordination} (institutional) and \emph{reward model} aspects, respectively. Related forms include \textit{political pluralism}, which is widely regarded as an important condition of democratic life, and \textit{legal pluralism}, which becomes especially salient where multiple legal orders, jurisdictions, or normative authorities overlap (e.g. jurisdiction in international matters).

All these forms of pluralism arise naturally when adopting an ecological view of social systems: they mirror the \emph{diversity} of aspects across social niches available within a society at a given point in time. This implies that rules can be understood and modelled only within the context of social roles (which are, in turn, associated with specific niches). Any effort to capture regularities, rules, or associated epiphenomena without considering the niche using them is bound to fail, because as soon as the relative distribution of niches in society changes, the foundations of the model will change as well. This observation is compatible with accounts emphasizing the \emph{material} dimension of law, according to which law is primarily constituted and performed out of dedicated practices pertaining to a certain social assemblage \citep{Pottage2012,Philippopoulos-Mihalopoulos2014,Gordon2022}. Conversely, from a system-design perspective, this view entails that all artifacts capturing norms (norm as in \emph{normativity}, but also as in \emph{normality}) need to be considered in an ecologically sound computational architecture, able to assimilate or accommodate adequately any environmental configuration, that is, capable of representing and handling concerns of norm applicability, interpretation, and revision. This requirement suggests introducing an explicit \emph{normware} design level for artificial devices, extending the distinction between hardware and software \citep{Sileno2018a,sileno2026}.

\section{Discussion}

\subsection{Institutional Metaphors for Large-Scale Distributed AI}

With the increasing use of distributed, computational, and possibly autonomous systems---such as those employed in Agentic AI---as well as technologies based on distributed ledgers/smart contracts, the Internet of Things, and digital twins, it is relevant to apply the preceding conceptualizations to assess the extent to which the institutional mechanisms identified in the legal domain find correspondence within computational social systems.

\subsubsection*{Agentic AI}
As observed throughout this chapter, the rapid deployment of Agentic AI presents an  operational testbed for institutional metaphors. In these systems, LLMs are wrapped within a software \textit{harness} or \textit{scaffolding} managing execution loops, memory buffers, external tool calls, and guardrail enforcement. System prompts and
guardrail policies may function as norms ($N$) when they guide or constrain conduct; shared records, authorised workflow states, or ledger entries may function as institutional facts ($F$) when they are treated as authoritative within the relevant arrangement. An agent harness, although generally specified in terms of \textit{workflow}, functions rather as an electronic institution, as it can be used to define operational boundaries, interaction protocols, and power distributions across heterogeneous actors. Contemporary architectures, however, frequently suffer from structural flaws derived from primitive organizational systems. 

First, many default workflows have used the same underlying LLM both to \emph{generate} a solution and to \emph{verify} it. As noted in Section~\ref{subsec:what-agents-look-like-externally}, effective verification depends on forms of independence, such as information asymmetry and distinct evaluative biases. When generator and verifier share identical parametric knowledge and context buffers, verification collapses into sycophancy or defensive rationalizations due to confirmation bias rather than critical oversight. Contemporary architectures increasingly address this failure mode by integrating heterogeneous pipelines including non-generative models---such as Jev-like discriminative models, JEPA-based world models, or formal theorem provers (Lean, Coq, etc.). 

Second, computer science traditionally focuses on \emph{ex-ante} enforcement mechanisms---hardcoding constraints through, for instance, schema validation, API tool-masking, or interceptors that inhibit agent affordances ($A$) prior to execution. This approach conflates deontic prohibitions (what an agent \emph{ought not} to do) with potestative disabilities (what an agent \emph{cannot} do). As multi-agent systems scale in complexity, however, verifying every step via assertion may become not only computationally prohibitive, but also \textit{operationally brittle}, or overly restrictive, as it may disable valid agent affordances under edge cases (see the discussion in the operating systems paragraph). 


Techniques like \textit{Reinforcement Learning from Human Feedback} (RLHF) have been introduced to accommodate contexts where rules cannot be easily hardcoded; yet, bottom-up mechanisms for preference aggregation fail to match the top-down authoritative mechanisms observable in legal systems. In this context, designing agent harnesses with the right combination of procedural mechanisms and case-centred contextualization points---such as explicit escalation roles and institutional appeal protocols---would allow higher-order processes and human overseers to evaluate reasons for failure and dynamically reallocate burdens of proof, echoing human legal systems.

\subsubsection*{Smart contracts}Despite their name, ``smart contracts'' (see, \emph{e.g.}, \cite{DeFilippi2020}) neither embody any problem-solving methodology nor are specified by assigning normative positions to parties as in conventional contracts. Their primary innovation, much like the blockchain, is the \emph{distributed ledger}, which is utilized for both ``contract'' publication and the registration of related transactions, thereby removing the requirement for an explicit third-party maintainer (e.g., a bank, public administration, or notary). Consequently, they lay the groundwork for a potential infrastructural centralization of registration power. However, by collapsing normative functions into the implementation layer, these artifacts remain in practice unintelligible to users. Furthermore, they fail to provide by design architectural mechanisms allowing interpretative power to exercise negative feedback on directive power in novel or exceptional contexts not anticipated at design time. This limitation severely undermines their suitability for institutional operationalization. 

\subsubsection*{Operating systems (at large)}
In recent years, many efforts have been directed toward the development of \emph{secure} operating systems. Traditionally, most implementations build upon solutions in the spirit of \emph{access control lists} (ACLs)---mapping each user and object to the set of operations they are permitted to perform (e.g., read, write, execute) \citep{Ferraiolo1999}. These permissions are also commonly referred to as privileges or authorizations; however, upon closer inspection, this represents a conflation of the homonymous Hohfeldian positions. Without such permission, a user is \emph{disabled} from performing the action rather than prohibited.\footnote{For a formal analysis of the interaction between deontic and potestative categories, see e.g. \citet{sileno2023}.}. By virtue of their dematerialized content, computational systems are not fundamentally different from institutional systems: both rely on symbol-processing mechanisms. In this sense, writing a file is not merely a physical operation but an institutional one; consequently, performing it requires the corresponding institutional power. Capability-based security models implicitly rely on this principle, utilizing communicable tokens for authorization known as \emph{capabilities}; see, e.g. \citep{Levy1984}. The \emph{principle of least privilege} (or least authorization) requires that capabilities be assigned strictly according to the actual operational requirements of the application. 

However, treating authorization solely as power introduces further complexities. In socio-legal settings, permission is most commonly invoked when an agent possesses the (typically physical) power to perform an action $A$ but is legally prohibited from doing so (e.g. you are permitted to smoke when it is generally prohibited to). More generally, permissions are introduced because power lacks the necessary granularity to capture outcomes and higher-order systemic effects, whereas prohibition can capture those. For instance, while a single action \emph{per se} may be permitted (enabled), that action within a sequence or in combination with others may yield unauthorized results (consider, for example, \emph{denial-of-service} (DoS) attacks exploiting trusted applications).\footnote{As an additional example of engineering ``blindness'' to social semantics, consider the endless discussions regarding which logical fragment should be adopted for knowledge representation on the Semantic Web---focusing primarily on computability and the halting problem---disregarding the fact that as soon as an external interface is opened via SPARQL, the system can be rendered inoperable by a DoS attack, regardless of the logical fragment employed.}

Failure cases extracted from \emph{ex-post} evaluations can be used effectively to monitor preparations for known failure schemes. However, the fundamental problem---addressed in human societies through deontic directives---is establishing \textit{principled reasons} as an institutional anchor to qualify \emph{new} types of failure (or to create new qualification rules) within an existing normative framework. Organizations heavily struggle to maintain alignment across their official AI policies, \textit{instructions for use} (IFU) for human operators, system prompts governing LLMs, instructions for monitoring agents, and hard access controls. As a result, they remain ill-prepared to integrate lessons learned in a timely manner or adapt their governance frameworks when unexpected edge cases arise. Clearly separating deontic and potestative dimensions is a first step necessary to overcome these difficulties. 

\subsubsection*{Reinforcement learning, and genetic algorithms}

In the last decade, major results have been obtained through the hybrid application of deep learning methods and reinforcement learning (starting from AlphaGo \citep{Silver2016}). Genetic algorithms address related optimisation problems, through evolutionary selection over candidate solutions. Most of these approaches depend on centralized evaluative mechanisms---reward functions, fitness functions, or preference models---that determine which outcomes count as better or worse. These types of feedback represent only two among the six primitive reward and punishment regimes identifiable in institutional systems (see \cite[sec.~9.5]{Sileno2016}), but, independently of their type, the deeper obstacle to generalising such mechanisms is the difficulty of specifying a general feedback function.

AI researchers and practitioners (and problem-solvers more generally) tend to assume that identifying goals is the easy part, whereas determining how to reach those targets is the hard part. However, when conceiving applications required to adapt to users or social environments, this presumption rapidly collapses---and necessarily so---for at least three key reasons: (a) the multiplicity of preference configurations among social participants, (b) the highly contextual nature of these configurations, and (c) the predominantly tacit character of human preferences.

To some extent, RLHF belongs to this same architectural lineage, although it aggregates feedback from diverse human annotators. This raises a fundamental underlying question: what is the specific operational niche to which this feedback is anchored? While majority voting or reward-model averaging can successfully isolate recurrent patterns across annotators, this aggregation mechanism risks collapsing the niche-alignment problem into alignment with merely the most statistically represented niche.

\subsubsection*{General critique and the role of AI \& Law}

A recurrent limitation of contemporary AI lies in trying to capitalize too much on optimizing and reasoning about evidential input and options for action (\emph{generate \& test} paradigm, or in more recent terminology, \emph{generate \& verify}) within known requirements, constraints, and problem formulations, rather than examining the underlying phenomena of niche construction, adaptiveness, and the derivation of requirements linked to social roles. Deciding to target social roles rather than formulating requirements on an individual basis follows from Heiner's theory and is cognitively supported by the observation that humans actually perform better at deontic reasoning than at evidential reasoning \citep{Mercier2011}, acquiring this capability earlier in childhood.

Interestingly, the AI \& Law discipline \citep{Bench-Capon2012} can be thought of as originating from reversing this general AI attitude---that is, by focusing strongly on conflicts between goals and the mechanisms by which we acquire and select them (\emph{normativity}). Capitalizing on this experience, even if considering the automatic discovery of ecological niches too ambitious, one could still aim for the automated design of compositions of social roles intended to achieve specific given requirements. The core problem in constructing such systems lies in detecting and resolving internal role conflicts by internalizing epistemological, normative, and value pluralism as a design issue rather than just a monitoring one. A critical design challenge involves establishing adequate data and knowledge roles---those that mediate data and knowledge flows between behavioral niches. An improvident centralization of data, for instance, might enable the utilization of collected information for purposes unintended by the data-providing agents. This reveals the need for a \emph{principle of \textit{data-information minimization}}, whereby the provision and collection of data are calibrated against the adequacy of reasons for accessing information (conditioned upon the social role attributed to the requestor and its accompanying social context), rather than relying on a token-based principle of least privilege tied solely to the requestor. In other words, this calls for a fundamental reconsideration of the attribution of responsibility to computational entities and social stakeholders (see, e.g., \cite{Sileno2020}).

\subsection{AI Techniques for Running Institutions} 

\subsubsection{Confronting inadequate simplifications in a complex world}

With globalization, all socio-economic actors (private individuals, private companies, NGOs, and public institutions) became increasingly active at an international scale, and institutional interdependencies have transitioned from exceptional to standard operating conditions. Operating concurrently within multiple legal jurisdictions, these actors are subject to an ever-changing web of rules (e.g., national provisions, international treaties, standards), whose complexity naturally increases with the volume of international activities undertaken. This carries risks of normative misalignment, which may expose such actors to enforcement penalties and institutions to systemic failures. On the other hand, as their activities occur at an international level, those actors can establish practices---for instance, aiming to reduce the burden of compliance with environmental, labour, or privacy regulations, or implementing schemes of tax avoidance, tax evasion, or money laundering---that are not yet recognized by---or remain opaque to--- formal institutions at the national level. 

Far from simplifying the operational environment, contemporary geopolitical disruptions are just increasing systemic complexity. Because of the deep transnational interdependence, political, socio-economic, and environmental shocks remain no longer isolated; and socio-economic actors find themselves in an increasingly turbulent environment.

From a theoretical perspective, an increase in complexity means an increase in the C-D gap for agents, and therefore creates pressure toward the establishment of simpler, more manageable operational rules---at the potential expense of the normative elaboration conducted upfront and of the diversity of cases that may occur, cases that would otherwise require a dedicated response to properly serve the interests of citizens. In this context, AI, and in particular AI \& Law, could provide methodological and computational support for consolidating the \emph{rational} functioning of collective agencies.

Beyond classic problem-solving tasks for operational purposes (e.g., \emph{planning}, \emph{scheduling}), here we refer in particular to methods supporting higher-order functions of rational agency as those highlighted above.  For instance, in current practices, one can observe little attention to pluralism within organizational settings, driven by an exclusive focus on the implementation and optimization of ``normal'' provision patterns---the so-called \emph{happy flow}. Consequently, even though the presumption of compliance utilized at the rule design phase is generally untested, patterns of failure are relegated outside the scope of attention. Thus, even if numerically negligible, failures in service delivery consume an increasingly disproportionate share of resources, thereby generating the well-documented phenomena associated with \emph{machine bureaucracy} (see, for instance, \cite{Seddon2008}).

Symptomatic of this situation are unrealistic methods for quantifying and measuring performance, a general disregard for the \emph{systemic} significance of failures, and the absence of systematic diagnostic processes capable of mitigating operational burdens (see,  e.g. \cite{Sileno2017}). Alongside this, excessive expectations regarding the insights derivable from big data analytics remain fundamentally implausible, precisely because the specific social niches in which data is generated are completely neglected. 

On the opposite side, collective agencies may risk adopting an attitude of \emph{paralyzing realism} when they pay excessive attention to capturing societal pluralism at an overly fine granularity, rendering them incapable of completing the requirements aggregation phase necessary to advance to the directive, operational, or regulative stages.

Both the machine-bureaucracy and paralyzing-realism scenarios stem from failing to establish an adequate systematization of environmental modeling---whether through under-modeling or over-modeling. The solution lies in building appropriate networks to distribute the burden of capturing social requirements and non-compliance scenarios, ensuring these insights are transmitted, maintained, and processed to support both the design and development phases and ongoing operational execution (e.g., through automated model-based interpretation and diagnostics \citep{Boer2011d, Boer2011b, Sileno2016, Sileno2017}).

These vulnerabilities are complementarily increased by the rapid entrenchment of contemporary AI tools---particularly Generative AI---into public administration and organizational operations at large (see e.g. \citep{Engstrom2020}). Organizations increasingly rely on proprietary, LLM-driven systems to synthesize evidence, draft policy options, evaluate claims, and interface with the public. This tendency shifts organizational agency toward external computational intermediaries, introducing significant structural dependencies. On a geopolitical scale, this exposes organizations to technological lock-in, extraterritorial regulatory shifts, and vulnerabilities in operational continuity. At cognitive and societal levels, uncritical adoption risks eroding institutional memory, flattening local context, and inducing systemic confirmation bias among human decision-makers \citep{Cobbe2021}. When administrative staff accept automated recommendations as baseline truths---a phenomenon driven by \textit{cognitive offloading} under high workloads---the deliberative capacity of the institution diminishes \citep{Goddard2012, alon-barkat2023}. Rather than enhancing administrative rationality, the uncritical integration of AI models risks entrenching existing biases and institutionalizing forms of bureaucratic alienation.

\subsubsection{Possible lines of action}

AI research, particularly within AI \& Law and related tracks, offers a constructive path forward by operationalizing these mechanisms into a computational architecture. 

First, exploiting case law and judicial precedents (or functionally similar domains, i.e. use/failure cases and precedent organizational decisions), computational frameworks for \textit{narrative} and \textit{argumentation analysis} can help organizations to identify respectively relevant sequences of events and reasons (e.g. factors for decisions). These can be further studied to track jurisdictional drift, and model the evolution of the normative systems, and will enable organizations to move beyond brittle \emph{ex-ante} rule codification, establishing principled reasons to qualify new failure modes as they arise in real-world practice.

Second, AI \& Law provides in principle conceptual tools to replace rigid, linear workflows with adaptive organizational architectures. For instance, by embedding operational mechanisms such as recourse, appeal, and overruling directly into multi-agent harnesses, computational systems may automatically trigger a post-hoc audit when a failure is detected, 
dynamically reallocating burdens of proof, to orchestrate the interpretation of that failure and possibly revisions of policy. In other words, rather than treating failures as external anomalies, incorporating these feedback mechanisms help organizations to integrate lessons learned across policy layers, system prompts, access controls, and so on.

Third, by leveraging appropriate computational representations, AI \& Law enables the construction of context-aware executable specifications of norms. Contemporary language technologies may reduce, though not eliminate, the traditional knowledge-acquisition bottleneck.
The resulting normative models allow statutory provisions, which are inherently open-textured in natural language, to be interpreted and operationalized within the specific operational niches where agents interact. When deployed as the governing infrastructure for NMAS, these frameworks regulate commitments ($C$), affordances ($A$), expectations ($E$), and susceptibilities ($S$), providing accountability by design and thus without reducing governance either to rigid ex-ante constraints or to unmonitored adaptation. Simulation provides a further opportunity. Organisations may use models to stress-test candidate contracts, policies, or organisational procedures against environmental shocks; evaluate risk profiles \emph{ex ante}; and identify where execution requires additional safeguards or dispute-resolution mechanisms.

\subsection{Converging Opportunities}

A central challenge in operationalizing AI within institutional contexts lies in establishing intelligible intervention points that guarantee meaningful human oversight without degrading efficiency and effectiveness. From a dual perspective, however, we can say that proper and robust institutional design requires preserving and strategically engineering ``\emph{constructive frictions}'': points at which decisions can be questioned, evidence inspected, reasons articulated, and authority
reallocated.

Despite the enthusiasm for large-scale data-driven analytics, as well as the seemingly general-purposiveness of LLM-centered tools, the qualitative depth of knowledge available to and applicable-by decision-makers within collective agencies cannot substantially improve on its own. Qualitative depth can be captured only through more principled modeling and simulation methods targeting \emph{social niches}, combined with knowledge maintenance, refinement, reconciliation, and alignment techniques. This approach directly reduces our C-D gap, thereby providing robust support for designing more \emph{intelligent} organizational rules, that is, rules backed by explicit reasons and better situated with respect to their operational environments.

Privacy provides an exemplary, instructive testbed for these considerations, as it requires modeling norms governing information use \emph{within} niches and flows \emph{between} niches (cf. \emph{contextual integrity}, \cite{Nissenbaum2009}). More importantly, however, in the last few years calls have emerged for \emph{AI governance} and \emph{AI assurance}, to ensure that not only privacy, but also fairness, explainability, and contestability interests of stakeholders are adequately protected in automated decision-making systems utilizing machine learning techniques. These requirements depend heavily on the material conditions of the specific social niche. Skin color, for instance, is functionally relevant when purchasing cosmetics, and gender matters in dating applications, whereas both characteristics are highly problematic in hiring decisions. While it is sometimes assumed that fairness and explainability can be implemented via technical AI methods, in operational practice these requirements are rarely \emph{implemented} with AI techniques at all. They are typically discovered through iterative risk assessment and defined by the burdens of proof dictated by the scenario. From this practical perspective, fairness can be reframed as a property of the input data--its representativeness (of population, of outcomes, of utility, ...), labeling, and selection. Explainability maps instead to a narrative obligation; first justifying why statistics-based decision-making is acceptable in this context at all, and then justifying why the specific decision is defensible. Technical fixes, moreover, tend to displace the compliance problem onto the technique itself rather than resolve it, and their appropriateness must in any case be weighed against e.g. privacy considerations, as they may inadvertently expose sensitive information (see e.g. \cite{Chang2020}).

Addressing privacy and AI governance requires modeling underlying norms, institutionalizing attention to system failures, and adopting a diagnostic perspective toward organizations and their institutional goals. This posture is increasingly reflected in risk-tiered regulatory frameworks such as the EU AI Act, which make documentation, human oversight, and iterative risk management central to compliance.

In light of the preceding conceptualizations, a natural starting point for this innovation is a sufficiently rich methodology for describing and reasoning about social roles, such as those captured by CAES and Hohfeldian positions (DCPL). 

Because normative positions can shape commitments, affordances,
expectations, and susceptibilities, CAES/DCPL descriptions provide a
framework for understanding role players as boundedly rational agents.
Consequently, they can be used to characterize the operational problems---such as modeling, design, planning, monitoring, diagnosis, and assessment (see, e.g. \cite{Breuker1994})--- that agents must solve within a given social role, along with the knowledge resources required to do so.

The greater the problem complexity (viewed through the lens of the C-D gap), the more difficult it becomes for an observer to verify whether an agent is acting as expected. To address this, the agent must articulate and defend its choices to other participants in the environment to demonstrate correct conduct. Evidently, an agent \emph{in a given role} is inherently biased in its information gathering by its CAES description requirements. Therefore, correctly assessing \emph{ex-post} whether the agent reached sound decisions often requires evaluating rationality at a higher level through the dialectical interplay between agents. A standard solution, compatible with Heiner's theory of predictable behavior, involves establishing increasingly rigorous \emph{burden of proof} protocols for justifying decisions. Conversely, when performance can be readily scored (such as in facial recognition tasks), the requisite depth of the articulated reasoning process remains minimal. This brings us to a fundamental design choice: deciding whether to employ \textit{tacit knowledge} methods (typically statistical and feed-forward) or \textit{explicit knowledge} methods. Naturally, the former are inherently faster than the latter; however, speed is not the only value at stake.

For Watson-like systems, the results of the question-answering ``race'' among competing agents are straightforward: the human arbiter or judge is presumed to know the answer, and there are typically no demands to explain \emph{why} that answer is correct. Along similar lines, consider a facial recognition application for unlocking a smartphone. By contrast, when a diagnostic system concludes that ``the patient has appendicitis,'' it is natural for the user to demand \emph{why}, expecting reasons that are intelligible and acceptable within that domain. Similarly, a public administration cannot (or at least ought not to) deny a parking permit or citizenship application without supplying explicit justification. The \emph{social role} required of intelligent agents across these two classes of applications is fundamentally distinct, and this divergence lies at the root of contemporary calls for \emph{explainable AI} \citep{Miller2019}. When AI systems assume socially consequential roles, the relevant question is not whether every internal computation can be exposed, but whether the system can provide the kinds of reasons, evidence, and review mechanisms that its institutional role requires.

To conclude, if we solve the social-role acquisition problem, we can solve the explainable AI problem (at least in its societal dimension), because we will be able to identify a higher-order burden of proof (and associated protocols) that can be used to distribute computation to specialized agents. This problem lies at the core of our difficulties in designing complex, adaptive AI systems, as illustrated by the confused use of knowledge structures throughout the history of AI. Moreover, solving the social-role acquisition problem will also improve the functioning of our institutions, providing a better environmental model with which to implement and test new institutional mechanisms, and through which to interpret social data.

Automating the social role acquisition problem from scratch would require the acquiring agent to be embedded in the social system in the same---or a very similar---way as humans are. This is currently not a realizable condition, for major technical and social obstacles. However, a first step toward solving the social-role acquisition problem is realizable: considering as primary design source the collective requirements as those that communities reify in institutions, and applying the lessons learned therefrom in engineering.

\printbibliography

\end{document}